\documentclass[conference]{IEEEtran}
\usepackage{graphicx}
\usepackage{algorithm}
\usepackage{algpseudocode}%
\usepackage{enumerate}

\usepackage{amsthm}
\makeatletter
\def\th@newremark{\th@remark\thm@headfont{\bfseries}}
\makeatletter
\usepackage[cmex10]{amsmath}
 {  
}

\theoremstyle{newremark}

\theoremstyle{newremark}
\newtheorem{rem}{Remark}

\theoremstyle{newremark}
\newtheorem{fact}{Fact}

\theoremstyle{newremark}

\theoremstyle{newremark}

\theoremstyle{newremark}

\theoremstyle{definition}

\theoremstyle{newremark}

\DeclareMathOperator*{\argmin}{arg\,min}
\usepackage{amssymb}

\begin{document}
\setlength{\abovedisplayskip}{3pt}
\setlength{\belowdisplayskip}{3pt}

\title{Detection for 5G-NOMA: An Online Adaptive Machine Learning Approach}
				
\author{\IEEEauthorblockN{Daniyal Amir Awan\IEEEauthorrefmark{1},
Renato L.G. Cavalcante\IEEEauthorrefmark{1}\IEEEauthorrefmark{2}, Masahiro Yukawa \IEEEauthorrefmark{3}\IEEEauthorrefmark{4} and 
Slawomir Stanczak\IEEEauthorrefmark{1}\IEEEauthorrefmark{2} 
\IEEEauthorblockA{\IEEEauthorrefmark{1}Technical University Of Berlin, Berlin, Germany, Einsteinufer 27, 10587 Berlin, Germany,}
\IEEEauthorblockA{\IEEEauthorrefmark{2}Fraunhofer Institute for Telecommunications, Heinrich Hertz Institute, Einsteinufer 37, 10587 Berlin, Germany,}
\IEEEauthorblockA{\IEEEauthorrefmark{3}Department of Electronics and Electrical Engineering, Keio University, Japan,}
\IEEEauthorblockA{\IEEEauthorrefmark{4}Center for Advanced Intelligence Project, RIKEN, Japan.}
Email: \{daniyal.a.awan, slawomir.stanczak\}@tu-berlin.de, renato.cavalcante@hhi.fraunhofer.de, yukawa@elec.keio.ac.jp}}

\maketitle

\begin{abstract}
Non-orthogonal multiple access (NOMA) has emerged as a promising radio access technique for enabling the performance enhancements promised by the fifth-generation (5G) networks in terms of connectivity, latency, and spectrum efficiency. In the NOMA uplink, successive interference cancellation (SIC) based detection with device clustering has been suggested. In the case of multiple receive antennas, SIC can be combined with the minimum mean-squared error (MMSE) beamforming. However, there exists a tradeoff between the NOMA cluster size and the incurred SIC error. Larger clusters lead to larger errors but they are desirable from the spectrum efficiency and connectivity point of view. We propose a novel online learning based detection for the NOMA uplink. In particular, we design an online adaptive filter in the sum space of linear and Gaussian reproducing kernel Hilbert spaces (RKHSs). Such a sum space design is robust against variations of a dynamic wireless network that can deteriorate the performance of a purely nonlinear adaptive filter. We demonstrate by simulations that the proposed method outperforms the MMSE-SIC based detection for large cluster sizes.     
\end{abstract}

\section{Introduction}\label{sec:introduction}
Non-orthogonal multiple access (NOMA)\footnote{In this study we use the term NOMA to refer to power-domain NOMA.} is being discussed in the context of fifth generation (5G) networks as one of the key technological enablers for systems such as the \textit{5G New Radio} (5G-NR) and the \textit{Internet of things} (IoT) \cite{Shin2017,Wang2016,Ding2017}. 5G NOMA promises to enable higher user density, lower latency, and higher spectrum efficiency when compared to fourth generation (4G) networks \cite{Wang2016}. 5G enabled IoT, for example, will comprise massive number of devices transmitting sporadically with diverse latency and data rate requirements. There is a reasonable consensus that massive machine-type-communications (mMTC) or IoT systems cannot be supported by the current LTE uplink due to scheduling delays and signaling overheads \cite{3GPP}. As a result, NOMA has become a subject of increasing interest in the wireless communication community \cite{IslamZD17,Shin2017}. 
		
		Orthogonal multiple access (OMA) systems, such as those based on orthogonal frequency-division multiple access (OFDMA), can only support a moderate number of active devices simultaneously due to a limited number of orthogonal frequency-time resource blocks (RBs). In these systems, only a fraction of the total system bandwidth is available to the scheduled devices. NOMA, in contrast, allows multiple devices to be superimposed, in principle, on the entire system bandwidth, and multiplexing is done in the power (or signal-to-noise ratio (SNR)) domain. Therefore, when compared to OMA systems, a significantly larger number of devices can transmit simultaneously and enjoy the entire system bandwidth. Intuitively, NOMA leads to improvements in active device density, latency, and spectral efficiency \cite{IslamZD17}. Indeed, from the spectral efficiency point of view, NOMA has been shown to be theoretically optimal both in the uplink and the downlink in a single-cell network \cite{Shin2017}. System level experiments have also shown promising results in terms of cell-throughput \cite{Benjebbour2015,3GPP}. 

     In the NOMA uplink, demultiplexing of devices is usually achieved by means of successive interference cancellation (SIC) at the base station (BS) \cite{Shin2017}. If multiple antennas are available at the BS, SIC can be combined with minimum mean-squared error (MMSE) beamforming \cite{HIGUCHI2015}. To maximize the spectral efficiency, all devices should be superimposed on the entire system bandwidth in a single-cell NOMA with SIC \cite{Shin2017}. The problem is that, for a large number of devices, cochannel interference seriously degrades the performance of MMSE-SIC in terms of SIC complexity and error propagation \cite{Tabassum2016,Ding2017,Shin2017}. There is therefore a strong need for an alternate way of implementing the NOMA uplink at the cost of lower spectral efficiency. A straight forward approach is to cluster devices, such that devices belonging to the same cluster transmit simultaneously on a set of orthogonal RBs. The sets assigned to each cluster are disjoint to avoid cochannel interference.
MMSE-SIC NOMA can then be performed within each cluster. However, the size of each cluster is restricted by the fixed number of BS antennas, SIC complexity, and SIC errors. This shows that there is an inherent tradeoff between the cluster size and the incurred SIC error \cite{Tabassum2016,Shin2017}, and new interference cancellation and detection approaches are needed to maximize the cluster size while keeping SIC errors to a low level. 
 
    The contribution of this study is a novel detection method for the NOMA uplink based on online learning. A similar approach was taken in \cite{Theodoridis2011, SlavakisSliding2008}, where the authors proposed online learning based beamforming in a \textit{reproducing kernel Hilbert space} (RKHS) associated with a Gaussian kernel. The strength of this nonlinear filtering technique in an RKHS lies in the possibility of mapping spatial signatures of devices to a higher dimensional space, where they can be better resolved as compared to the original space. Furthermore, the adaptive filtering algorithm requires simple computations. Indeed, the authors showed that this high-resolution nonlinear filter exhibits better interference and noise cancellation capability as compared to a linear filter, especially when the number of antennas at the BS is less than the number of active devices. Nonlinear beamformers are however highly sensitive to variations in the environment and the performance may deteriorate in dynamic networks, such as IoT networks, where devices may join the network sporadically. Therefore, in contrast to the nonlinear beamforming filter proposed in \cite{Theodoridis2011, SlavakisSliding2008}, we propose a partially linear beamforming filter design in the sum space (RKHS) of a linear and a nonlinear RKHS. This sum space adaptive filter enjoys high resolution and exhibits robustness against variations in the environment. Simulations show that our method outperforms both the MMSE-SIC based detection and the nonlinear adaptive filtering in a dynamic wireless environment. 

\section{Preliminaries}\label{sec:preliminaries}
The sets of real numbers, non-negative integers, positive integers, and complex numbers are denoted by $\mathbb{R}$, $\mathbb{Z}_{\geq 0}$, $\mathbb{Z}_{>0}$, and $\mathbb{C}$, respectively. 
We denote by $\text{span}\left\{\mathcal{V}\right\}$ the set comprising all finite linear combinations of the elements of $\mathcal{V}$. We define $\overline{N_{1},N_{2}}:=\left\{N_1,N_{1}+1,\ldots,N_2\right\}$, where $N_1, N_2 \in \mathbb{Z}_{\geq 0}$ with $N_1\leq N_2$. 

In this study, we deal with the problem of function estimation in special function spaces called \textit{Reproducing Kernel Hilbert Spaces} (RKHSs). In the following we briefly revisit selected aspects of the RKHS theory \cite{Scholkopf2001}. 
\subsection{Reproducing Kernel Hilbert Spaces}\label{sec:reproducing_hilbert_kernel_space}
Given an arbitrary set $\mathcal{U}\subseteq \mathbb{R}^{l}$, a function $\kappa:\mathcal{U}\times\mathcal{U}\rightarrow \mathbb{R}$ is said to be a kernel if it satisfies the following properties:
\begin{enumerate}
\item (\textit{Symmetry}) $\kappa(\mathbf{u},\mathbf{v})=\kappa(\mathbf{v},\mathbf{u}).$
\item (\textit{Non-negativity}) $\forall N \geq 1$, $\forall (\alpha_1,\alpha_2,\ldots,\alpha_N)\in\mathbb{R}^N$ and $\forall (\mathbf{u}_1,\mathbf{u}_2,\ldots,\mathbf{u}_N)\in\mathcal{U}^N$,  $\sum^{N}_{i=1}\sum^{N}_{j=1}\alpha_i\alpha_j\kappa(\mathbf{u}_i,\mathbf{u}_j)\geq 0$.
 \end{enumerate}
In the remainder of this study, $\kappa$ is either the linear kernel denoted by $\kappa_{L}$ and defined as $\kappa_{L}(\mathbf{u},\mathbf{v}):=\mathbf{u}^{T}\mathbf{v}$ or the Gaussian kernel denoted by $\kappa_{G}$ and defined as $ \kappa_{G}(\mathbf{u},\mathbf{v}):=\exp\left(\frac{\left\|\mathbf{u}-\mathbf{v}\right\|^{2}_{\mathbb{R}}}{2\sigma^{2}}\right)$.
Given a kernel $\kappa$, consider the linear space $\mathcal{H}_0$ of functions given by $f\in\text{span}\left\{\kappa(\mathbf{u},\cdot):\mathbf{u}\in\mathcal{U}\right\}$. Let the scalar multiplication and function addition in  $\mathcal{H}_0$ be defined in the usual way. For two functions $f:=\sum_{n=1}^{N}a_n\kappa(\mathbf{u}_n,\cdot)$ and $g:=\sum_{m=1}^{M}b_m\kappa(\mathbf{v}_m,\cdot)$, where $a_m,b_m\in\mathbb{R}$ and $\mathbf{u}_n,\mathbf{v}_m \in \mathcal{U}$, we define the inner product to be
\begin{equation}
       \left\langle f,g\right\rangle_{\mathcal{H}_0}:= \sum_{n=1}^{N}\sum_{m=1}^{M}a_n b_m\kappa(\mathbf{u}_n,\mathbf{v}_m),
    \label{eqn:inner_product_definition}
\end{equation}
with the induced norm given by
\begin{equation}
             \left\|f\right\|^{2}_{\mathcal{H}_o}=\left\langle f,f\right\rangle_{\mathcal{H}_o}.      
\end{equation}
The space $\mathcal{H}_{o}$ equipped with the above inner product is a pre-Hilbert space. We complete this space by including all limit points to obtain the reproducing kernel Hilbert space (RKHS) $\mathcal{H}$ uniquely associated with the kernel $\kappa$. In this study, we exploit the following two properties of the RKHS $\mathcal{H}$ associated with $\kappa$: 
\begin{enumerate}
	\item (\textit{Representation Property}) $(\forall\mathbf{u}\in\mathcal{U})$ $\kappa(\mathbf{u},\cdot)\in\mathcal{H}$.
   \item (\textit{Reproducing Property}) $(\forall f \in\mathcal{H})$ $(\forall\mathbf{u}\in\mathcal{U})$ $f(\mathbf{u})=\left\langle f,\kappa(\mathbf{u},\cdot)\right\rangle_{\mathcal{H}}$, where the inner product $\left\langle\cdot,\cdot\right\rangle_{\mathcal{H}}$ is the extension of the inner product $\left\langle\cdot,\cdot\right\rangle_{\mathcal{H}_{o}}$, defined in \eqref{eqn:inner_product_definition}, to $\mathcal{H}$ consisting of simple kernel evaluations. 
\end{enumerate}

\section{Partially Linear Adaptive Filtering Model for NOMA}
In this section, we describe the underlying adaptive filtering model for a single-cell NOMA uplink, which is inspired by the adaptive beamforming filter model in \cite{Slavakis2009}. A number of single antenna devices can transmit their data to a base station (BS). Following the approach in recent NOMA studies \cite{Tabassum2016,Shin2017}, devices in the cell can be divided in \textit{K-}device clusters with $K>1$, and each cluster is allocated a set of orthogonal resource blocks (RBs). Since the sets assigned to each cluster are disjoint, there is no inter-cluster interference. In the remainder of this study, we describe the adaptive filtering model for a single NOMA cluster which can be applied to all the clusters in the cell. 

We assume that the BS is equipped with a uniform linear array (ULA)\footnote{We consider a ULA for simplicity, but the proposed method can be extended to other configurations.} consisting of $M\in\mathbb{Z}_{>0}$ antennas. We assume a non-dispersive channel so that the received signal (sampled at a fixed symbol rate) is given by \cite{Slavakis2009}, $\forall k\in \overline{1,K}$,
\begin{align}
	\mathbb{Z}_{\geq 0}\rightarrow \mathbb{C}^{M}:\mathbf{r}(t)&:=\left[{r}_{1}(t),{r}_{2}(t),\ldots,{r}_{M}(t)\right]^{\intercal}\\
		           &:=\sum_{k=1}^{K}\sqrt{p_{k}(t)}h_{k}(t)b_{k}(t)\mathbf{s}_{k}(t) + \mathbf{n}(t),\nonumber
		\label{eqn:uplink_signal}
\end{align} 
where $h_{k}(t) \in\mathbb{C}$ is the channel gain and $b_{k}(t)\in\mathbb{C}$ is the modulation symbol (e.g., a BPSK or QAM symbol) for the $t$th sample. The vectors $\mathbf{s}_{k}(t)\in\mathbb{C}^M$ and $\mathbf{n}(t)\in\mathbb{C}^M$ stand for the array spatial signature of device $k$ and additive independent circularly-symmetric noise, respectively.

In this study, the objective is to design an online adaptive (beamforming) filter $g^{k}:\mathbb{C}^{M}\rightarrow\mathbb{C}$ for each device $k\in\overline{1,K}$, such that $(\forall t\in\mathbb{Z}_{\geq0})$ $\left|g^{k}(\mathbf{r}(t))-{b}_{k}(t)\right|\leq\epsilon$, where $\epsilon>0$ is a small predefined noise tolerance. In the following we omit the index $k$ for notational simplicity. 

\subsection{Adaptive Filtering In Sum Space RKHS}\label{sec:adaptive_filtering_in_sum_space}
The beamformer is typically a linear spatial filter of the form $g(\mathbf{r}(t))=\mathbf{w}^{H}\mathbf{r}(t)$, where $\mathbf{w}\in\mathbb{C}^{M}$ is chosen in such a way that the antenna beam is focused in the direction of the device of interest and the interference from other devices/directions is suppressed. A prominent example of a linear filter is the \textit{minimum mean-square error} (MMSE) beamformer. It is well-known that such a linear filter has only $M-1$ degrees of freedom (DOF) and it can typically resolve $K\leq M$ devices with sufficiently diverse spatial signatures \cite{Van2002}. Therefore, in a massive connectivity scenario, a system designer is currently forced to increase the number of antenna array elements $M$ to support a large number of concurrent devices. In contrast, nonlinear filters offer additional DOFs as well as enhanced interference and noise cancellation capabilities. In \cite{Slavakis2009}, the problem of nonlinear beamforming is addressed by employing the powerful RKHS framework associated with a Gaussian kernel. In more detail, the authors first show that there exists a bijection between complex and real vectors such that the task of designing the filter $g:\mathbb{C}^{M}\rightarrow\mathbb{C}$ can be transformed to the task of designing a filter $f:\mathbb{R}^{2M}\rightarrow\mathbb{R}$. The filter is then assumed to belong to a real infinite dimensional RKHS associated with a Gaussian kernel. 
The computational advantage of this approach is that the nonlinear filtering task in the original space becomes a linear filtering task in an infinite dimensional RKHS. The inner products are computed by simple kernel evaluations by the \textit{reproducing property}. Furthermore, spatial signatures of devices can be better resolved when mapped to a higher dimensional space by the \textit{representation property}. Indeed, it was shown in \cite{Slavakis2009} that this high resolution filter exhibits superior performance when $K>M$ as compared to a linear filter. On the other hand,  nonlinear spatial filters are highly sensitive to variations in the environment. For example, unlike linear filters, the filter response for a particular device $k\in\overline{1,K}$ in a NOMA cluster may deteriorate if one of the other devices $j\neq k$ becomes inactive. This behavior is clearly not desirable in dynamic wireless systems where devices transmit sporadically as, for example, in Internet of things (IoT) systems.  

To exploit the benefits of both linear and nonlinear spatial filters, we propose a partially linear filter in the sum space of the linear and the Gaussian RKHSs. In more detail, we denote by $\mathcal{H}_L$ and $\mathcal{H}_G$ the real RKHSs associated with the linear and the Gaussian kernel, respectively. The partially linear filter is defined as an element of the real RKHS $\mathcal{H}:=\mathcal{H}_L + \mathcal{H}_G:=\left\{w_{L}f_L+w_{G}f_G : f_L\in\mathcal{H}_{L}, f_G\in\mathcal{H}_{G}\right\}$, where $w_L,w_G >0$ are some weights for the linear and the Gaussian part, respectively. In this RKHS, the reproducing kernel and the inner product of $\mathcal{H}$ can be computed as follows: 
\begin{fact}[\textit{Reproducing kernel of the weighted sum space \cite{Yukawa2015}}]
\textit{Assume that the input space $\mathcal{U}\subseteq\mathbb{R}^{l}$ has a nonempty interior. Then, given any $w_{L},w_{G}>0$ and $\mathbf{u},\mathbf{v}\in\mathcal{U}$,  $\kappa(\mathbf{u},\mathbf{v}):=w_L\kappa_{\text{L}}(\mathbf{u},\mathbf{v})+w_G\kappa_{\text{G}}(\mathbf{u},\mathbf{v})$ is the reproducing kernel of the sum space $\mathcal{H}$ equipped with the inner product
  \begin{equation}
   \left\langle f,g\right\rangle_{\mathcal{H}}:=w_{L}^{-1}\left\langle f_L,g_L\right\rangle_{\mathcal{H}_L}+
                                                        w_{G}^{-1}\left\langle f_G,g_G\right\rangle_{\mathcal{H}_G}.
  \end{equation}
 }
\label{fact:mf}
\end{fact}
We demonstrate in Section \ref{sec:results} that we are able to enjoy the robustness of a linear design and the resolution of a nonlinear design by operating in the sum space of $\mathcal{H}_L$ and $\mathcal{H}_G$. We assume $w_L=w_G=1$ in the remainder of this section and Section \ref{sec:onlinea_filtering_algorithm}.
\subsection{The Online Adaptive Learning Problem}
\begin{figure}[t]
\centering
    \includegraphics[width=0.5\textwidth]{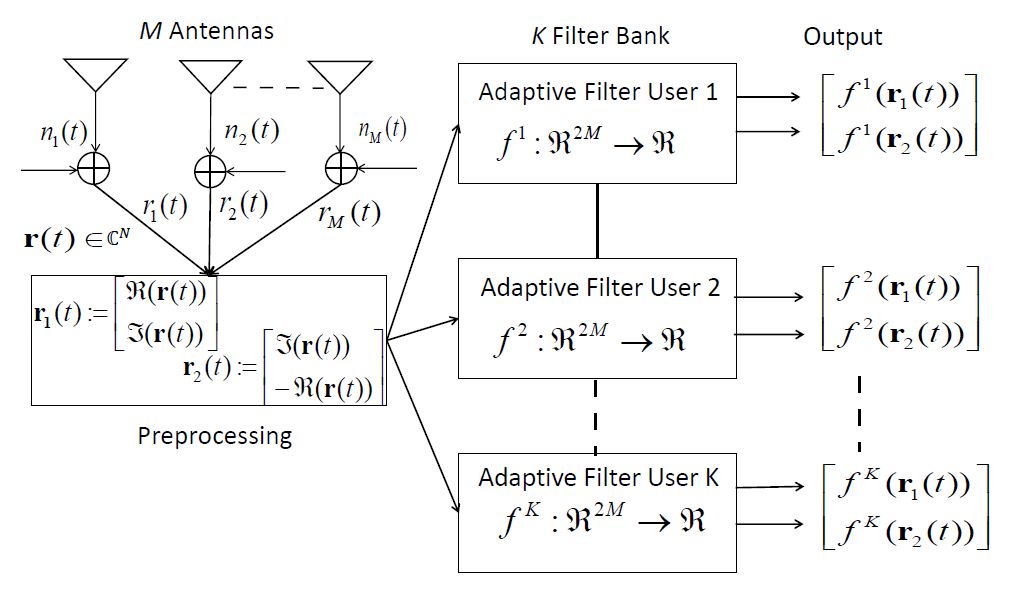}
		\caption{Adaptive beamforming/filtering model for a single cluster with K devices and a ULA with M antennas}
		\label{fig:beamforming_model}
\end{figure}

First, we convert the complex vector $\mathbf{r}(t)\in\mathbb{C}^{M}$ into two real vectors $\mathbf{r}_{1}(t):=\left[\Re(\mathbf{r}(t))^{\intercal}\,\Im(\mathbf{r}(t))^{\intercal}\right]^{\intercal}\in\mathbb{R}^{2M}$ and $\mathbf{r}_{2}(t):=\left[\Im(\mathbf{r}(t))^{\intercal}\, -\Re(\mathbf{r}(t))^{\intercal}\right]^{\intercal}\in\mathbb{R}^{2M}$ which enables processing in real Hilbert spaces as considered in \cite{Yamada2003,Slavakis2009}. Similarly, the training modulation symbols are converted to $\left[b_{1}(t)\,b_{2}(t)\right]^{\intercal}:=\left[\Re(b(t)) \,\Im(b(t))\right]^{\intercal} \in \mathbb{R}^2$. The proposed filter $f:\mathbb{R}^{2M}\rightarrow\mathbb{R}$ operates on $\mathbf{r}_{1}(t)$ and $\mathbf{r}_{2}(t)$ separately (as depicted in Fig. \ref{fig:beamforming_model}). The relation between $f$ and the complex valued filter $g$ described in Section \ref{sec:adaptive_filtering_in_sum_space} is given by $(\forall t\in\mathbb{Z}_{\geq 0})$ $\mathbb{C} \ni g(\mathbf{r}(t))=f(\mathbf{r}_{1}(t))+if(\mathbf{r}_{2}(t))$, where $i$ is the solution to the equation $i^2=-1$. To simplify indexing, we define a new time index $n:=2t+l-1$, $\mathbf{r}_{n}=\mathbf{r}_{2t+l-1}:=\mathbf{r}_{l}(t)$ and $b_{n}=b_{2t+l-1}:=b_{l}(t)$, $\forall t\in\mathbb{Z}_{\geq0}, \forall{l}\in\overline{1,2}$. Henceforth, we denote the input space of received signals by $\mathcal{U}:=\left\{\mathbf{r}_n\in\mathbb{R}^{2M}:n\in\mathbb{Z}_{\geq0}\right\}$.

We now turn our attention to the design of an adaptive filter $f$ such that $(\forall n\in\mathbb{Z}_{\geq 0})$ $|f(\mathbf{r}_n)-b_n|\leq \epsilon$, where the precision is controlled by the design parameter $\epsilon>0$. We assume that $f\in\mathcal{H}$ and a training sample $(\mathbf{r}_n,b_n)\in\mathcal{U} \times \mathbb{R}$ is available $\forall n\in\mathbb{Z}_{\geq 0}$. Then, a closed and convex set of functions in $\mathcal{H}$ consistent with the training sample at time $n$ is given by 
\begin{equation}
    C_n:=\left\{f\in\mathcal{H}:|\left\langle f,\kappa(\mathbf{r}_n,\cdot)\right\rangle_{\mathcal{H}}-b_n|\leq \epsilon\right\}.  
	\label{eqn:convex_set}
\end{equation}

In the online learning setting considered here, the training samples arrive sequentially and each sample defines a set of the form \eqref{eqn:convex_set}. Ideally, the objective is to find a filter $f^{\ast}\in\mathcal{H}$ such that $f^{\ast}$ is a member of all these sets, i.e., $f^{\ast}\in \bigcap_{n\in\mathbb{Z}_{\geq 0}}C_n$ . 
However, since it is challenging to find a low-complexity algorithm to solve this problem, we allow a finite number of sets not to share a common intersection and consider a simplified problem:
\begin{equation}
    \text{find}\, f^{\ast} \in \bigcap_{n\geq n_o}C_n,
	 \label{eqn:optimization problem}
\end{equation}
for some $n_o\in\mathbb{Z}_{\geq 0}$, under the assumption that $\bigcap_{n\geq n_o}C_n \neq \emptyset$. The advantage of \eqref{eqn:optimization problem} is that we can find an $f\in\mathcal{H}$ that is arbitrarily close to the intersection in \eqref{eqn:optimization problem} by means of the \textit{adaptive projected subgradient method} (APSM) \cite{Yamada2003,Theodoridis2011} which we describe below. 

As in \cite{Slavakis2009,Theodoridis2011}, given an index set $\mathcal{J}_n$, and starting from $f_0=0$, we construct a sequence of filter estimates in the sum space RKHS $\mathcal{H}$ (which stands in contrast to \cite{Slavakis2009,Theodoridis2011}) given as 
\begin{equation}
(\forall n\in\mathbb{Z}_\geq0)\, f_{n+1}=f_{n}+\left(\sum_{j\in\mathcal{J}_n}q^{n}_j\mathbf{P}_{C_j}(f_n)-f_n\right),
\label{eqn:apsm}
\end{equation}
where $\mathbf{P}_{C_j}(f_n):=\argmin_{f\in C_j} \|f_n-f\|_{\mathcal{H}}$ is the orthogonal projection of $f_{n}$ onto the set $C_j$, weighted by ${q}^{n}_{j}\geq 0$, $\forall j \in \mathcal{J}_{n}$ such that $\sum_{j\in\mathcal{J}_n}q^{n}_j=1$. The index set $\mathcal{J}_n$ allows for a subset of sets $C_1,C_2\ldots,C_n$ to be processed concurrently to improve the performance and the weights ${q}^{n}_{j}$ can be used to adaptively prioritize the sets. The computational advantage of this algorithm is that the projection $\mathbf{P}_{C_j}(f_n)$ only requires simple inner product operations \cite{Theodoridis2011}. 

Under certain assumptions, the sequence of estimates $(f_n)_{n\in\mathbb{Z}_{\geq 0}}$ converges to a point $f\in\mathcal{H}$ arbitrarily close to the intersection in \eqref{eqn:optimization problem} \cite{Yamada2003}. 
However, there are two practical issues with regards to the general form in \eqref{eqn:apsm}. 
The first one is a low-complexity construction of an index set $\mathcal{J}_{n}$, whereas the second issue is the computational complexity of iteration \eqref{eqn:apsm}. In more detail, it can be shown that the filter estimate generated by iteration \eqref{eqn:apsm} is given by the kernel series expansion $(\forall n\in\mathbb{Z}_{>0})$ $f_{n}=\sum_{i=0}^{n-1}\gamma^{(n)}_{i}\kappa(\mathbf{r}_i,\cdot)$ \cite{Theodoridis2011}. Therefore to keep track of $f_{n}$, the training samples $\mathbf{r}_i$ up to time $n-1$ have to be kept in the memory. Furthermore, the coefficients $\gamma^{(n)}_{i}$ are determined at each $n$ by the projection $\mathbf{P}_{C_j}(f_n)$ in \eqref{eqn:apsm}. Therefore, the memory requirements and the computational complexity may become prohibitive for large $n$. In the next section we provide low-complexity techniques to address these issues. 

\section{Online Adaptive Filtering Algorithm}\label{sec:onlinea_filtering_algorithm}
In this section we describe the steps of the proposed adaptive learning algorithm shown in Algorithm $1$. The \textit{Sample Update Step} and the \textit{Dictionary Update Step} of Algorithm $1$ address the two issues discussed above. The \textit{Adaptive Learning Step} is based on iteration \eqref{eqn:apsm}.

\subsection{Sample Update Step}\label{sec:sample_update}
We assume that at time $n\in\mathbb{Z}_{\geq 0}$ a training set $\{(\mathbf{r}_j,b_j):j\in\mathcal{J}_{n}, \mathbf{r}_j \in \mathcal{U}, b_j \in \mathbb{R}\}$ is available at the BS. A natural choice for $\mathcal{J}_{n}$ in a dynamic online setting is a small window $W\in\mathbb{Z}_{>0}$ of the most recent samples. To be more precise, we define $\mathcal{J}_{n}$ as $\mathcal{J}_{n}:=\overline{n-W+1,n}$ if $n \geq W-1$, otherwise $\mathcal{J}_{n}:=\overline{0,n}$. The window size $W$ is a design parameter chosen based on the available computational power. Larger sizes typically improve the performance at the cost of increased computational power.  
\subsection{Dictionary Update Step}\label{sec:dictionary_construction_step}
Recall that the filter estimate generated by \eqref{eqn:apsm} is given by $f_{n}:=\sum_{i=0}^{n-1}\gamma^{(n)}_{i}\kappa(\mathbf{r}_i,\cdot)$. This can be written as $f_{n}:=f_{L,n}+f_{G,n}=\sum_{i=0}^{n-1}\gamma^{(n)}_{i}\kappa_{L}(\mathbf{r}_i,\cdot)+\sum_{i=0}^{n-1}\gamma^{(n)}_{i}\kappa_{G}(\mathbf{r}_i,\cdot)$, where $f_{L,n}\in\mathcal{H}_L$ and $f_{G,n}\in\mathcal{H}_{G}$. By reducing the number of kernel terms in  $f_{L,n}$ and $f_{G,n}$ whose coefficients need to be updated at each iteration, the complexity of \eqref{eqn:apsm} can be reduced. Besides reducing the complexity, this \textit{online sparsification} also improves the predictive ability of the filter \cite{Engel2004}. To this end, instead of adding the most recent  $\kappa_{L}(\mathbf{r}_n,\cdot)$ and $\kappa_{G}(\mathbf{r}_n,\cdot)$ directly to the estimate $f_{n}$, we perform an admission control. The admission control checks if $\kappa_{L}(\mathbf{r}_n,\cdot)$ and $\kappa_{G}(\mathbf{r}_n,\cdot)$ can be approximated by a linear combination of kernel elements already admitted up to time $n-1$. The newly arriving elements are only added to the expansion $f_{n}$ if such an approximation is not possible. These so-called novel admitted elements, also known as a \textit{dictionary}, are kept in the memory and only their coefficients are updated in \textit{Adaptive Learning Step} based on iteration \eqref{eqn:apsm}.
\subsubsection{Dictionary for the linear part}
Since the RKHS $\mathcal{H}_{L}$ is the Euclidean space $\mathcal{U}\subseteq\mathbb{R}^{2M}$, every new element $\kappa_{L}(\mathbf{r}_n,\cdot)$ can be written in terms of a linear combination $\sum_{m=1}^{2M}[\mathbf{r}_n]_{m}\kappa_{L}(\mathbf{e}_m,\cdot)$ of the Euclidean basis $\mathcal{D}_{L}:=\left\{\kappa_{L}(\mathbf{e}_1,\cdot), \kappa_{L}(\mathbf{e}_2,\cdot),\ldots,\kappa_{L}(\mathbf{e}_{2M},\cdot)\right\}$, where $\mathbf{e}_m\in\mathbb{R}^{2M}$ is a vector having a one at the $m$th index and zeros elsewhere, and $[\mathbf{r}_n]_{m}$ is the $m$th entry of $\mathbf{r}_n$. 
As a result, it can be verified that $(\forall n\in\mathbb{Z}_{\geq 0})$ $f_{L,n}=\sum_{m=1}^{2M}\gamma^{(L,n)}_{m}\kappa_{L}(\mathbf{e}_m,\cdot)$ consists of only $2M$ basis kernels with coefficients $\gamma^{(L,n)}_{m}$ adapted at each iteration of the \textit{Adaptive Learning Step}. 
\subsubsection{Dictionary for the Gaussian part}
The dictionary $\mathcal{D}_{G,n}$ for the Gaussian part is adapted at each $n\in\mathbb{Z}_{\geq 0}$ by the following procedure \cite{Engel2004,SlavakisSliding2008}: 
\begin{enumerate}
 \item If $n=0$, the element $\kappa_{G}(\mathbf{r}_0,\cdot)$ is added, so that $\mathcal{D}_{G,0}:=\kappa_{G}(\mathbf{r}_0,\cdot)$.
 \item If $n>0$, the dictionary $\mathcal{D}_{G,n-1}$ is updated by $\kappa_{G}(\mathbf{r}_{n},\cdot)$ only if $\kappa_{G}(\mathbf{r}_{n},\cdot)$ is sufficiently novel in the sense explained below.
\end{enumerate}    

We denote by $\mathcal{H}_{G,n-1}:=\text{span}\{\mathcal{D}_{G,n-1}\}$ a closed subspace of $\mathcal{H}_{G}$. The novelty of the element $\kappa_{G}(\mathbf{r}_{n},\cdot)$ can be estimated based on its (approximate) linear independence, measured by a design parameter $\alpha>0$, from $\mathcal{D}_{G,n-1}$. The distance of $\kappa_{G}(\mathbf{r}_{n},\cdot)$ from $\mathcal{D}_{G,n-1}$ is given by $d_{n}:=\left\|\kappa_{G}(\mathbf{r}_{n},\cdot)-\mathbf{P}_{\mathcal{H}_{G,n-1}}(\kappa_{G}(\mathbf{r}_{n},\cdot))\right\|_{\mathcal{H}}^{2}$, where $\mathbf{P}_{\mathcal{H}_{G,n-1}}(\kappa_{G}(\mathbf{r}_{n},\cdot))$ is the orthogonal projection of $\kappa_{G}(\mathbf{r}_{n},\cdot)\in\mathcal{H}_{G}$ onto $\mathcal{H}_{G,n-1}$. If $d_{n}>\alpha$, the dictionary is updated by $\kappa_{G}(\mathbf{r}_{n},\cdot)$, i.e., $\mathcal{D}_{G,n}:=\mathcal{D}_{G,n-1}\cup \kappa_{G}(\mathbf{r}_{n},\cdot)$, otherwise there is no update and we have $\mathcal{D}_{G,n}:=\mathcal{D}_{G,n-1}$. The quantities $\mathbf{P}_{\mathcal{H}_{G,n-1}}(\kappa_{G}(\mathbf{r}_{n},\cdot))$ and $d_{n}$ are well defined and calculated as in Appendix \ref{app:A}.

In the following, we denote by $\mathcal{H}_{n}:=\mathcal{H}_L+\mathcal{H}_{G,n}$, where $\mathcal{H}_L:=\text{span}\{\mathcal{D}_{L}\}$ and $\mathcal{H}_{G,n}:=\text{span}\{\mathcal{D}_{G,n}\}$, closed subspaces of the sum space $\mathcal{H}$. Note that $\mathcal{H}_{G,n-1}\subseteq\mathcal{H}_{G,n}$ and $\mathcal{H}_{n-1}\subseteq\mathcal{H}_{n}$. 
\begin{rem}[Dictionary Pruning]\label{rem:memory_management}
In an online setting, a common approach to memory management is to restrict the overall dictionary size to some number $S\in\mathbb{Z}_{>0}$ by simply discarding outdated dictionary elements. In Section \ref{sec:results} we provide a pruning approach which is a natural candidate for a dynamic wireless network.    
\end{rem}
 
\begin{algorithm}[t]
\small
\caption{Online Adaptive Filtering Algorithm} 
\label{algorithm:one}
 \begin{algorithmic}
  \State \textbf{Initialization}: Fix $\epsilon>0$, training block length $T\in\mathbb{Z}_{>0}$, $W\in\mathbb{Z}_{>0}$, $\alpha>0$, $\mathcal{D}_{G,-1}:=\emptyset$, and $f_0=0$. 
 \State \textbf{At $n\geq0$}: 
\begin{enumerate}
  \item \textbf{Sample Update}: The training samples $\left\{(\mathbf{r}_j,b_j):j\in\mathcal{J}_{n}\right\}$ are available. Set $q^{n}_{j}=1/|\mathcal{J}_{n}|$, $\forall j \in \mathcal{J}_{n}$, where $|\mathcal{J}_{n}|$ is the cardinality of $\mathcal{J}_{n}$. 
  \item \textbf{Dictionary Update}: Follow the procedure in Section \ref{sec:dictionary_construction_step} to update the dictionary. 
	\item \textbf{Adaptive Learning}: Follow the procedure in Section \ref{sec:adaptive_learning_step} to calculate $f_{n+1}$.
	\end{enumerate}
	\textbf{If} $n=2T-1$ stop, \textbf{otherwise} go to Step 1.
 \end{algorithmic}
\end{algorithm}
\subsection{Adaptive Learning Step}\label{sec:adaptive_learning_step}
After the admission control in the \textit{Dictionary Update Step}, the projection $\mathbf{P}_{\mathcal{C}_{j}}(f_n)$ in \eqref{eqn:apsm} is given by the quantity $\beta^{n}_{j}\mathbf{P}_{\mathcal{H}_{n}}(\kappa(\mathbf{r}_{j},\cdot))+f_n$. The online learning iteration is given as, $\forall n \in \mathbb{Z}_{\geq 0}$,
\begin{align}\label{eqn:adaptive_learning_step}
\mathcal{H}_{n}\ni f_{n+1}:=f_{n}+\sum_{j\in\mathcal{J}_{n}}q^{n}_{j}\beta^{n}_{j}\mathbf{P}_{\mathcal{H}_{n}}(\kappa(\mathbf{r}_{j},\cdot)), 
\end{align}
where $\sum_{j\in\mathcal{J}_{n}}q^{n}_{j}=1$ and $\mathbf{P}_{\mathcal{H}_{n}}(\kappa(\mathbf{r}_{j},\cdot))=\sum_{m=1}^{2M}[\mathbf{r}_j]_{m}\kappa_{L}(\mathbf{e}_m,\cdot)+\mathbf{P}_{\mathcal{H}_{G,n}}(\kappa(\mathbf{r}_{j},\cdot))$ \cite{Yukawa2015}. The quantities $\mathbf{P}_{\mathcal{H}_{G,n}}(\kappa(\mathbf{r}_{j},\cdot))$ and $\beta^{n}_{j}$ are calculated as in Appendix \ref{app:B} and Appendix \ref{app:C}, respectively.
\begin{rem}[Complexity of Algorithm $1$]
The complexity of the \textit{Adaptive Learning Step} is linear in the dictionary size. The \textit{Dictionary Update Step} is the main computational load of Algorithm \ref{algorithm:one}. However, this complexity is upper bounded by $\mathcal{O}(S^2)$, where the constant $S$ is the upper bound on the size of the dictionary (See Remark \ref{rem:memory_management}). 
\end{rem}
 \section{Simulated Performance}\label{sec:results}
In this section we demonstrate the performance of our partially linear adaptive filter (PLAF) design by comparing its performance with a MMSE-SIC receiver and the purely nonlinear adaptive filter (NLAF) (see the discussion in Section \ref{sec:adaptive_filtering_in_sum_space}). For simplicity, we only consider intra-cluster (intra-cell) interference but Algorithm $1$ is also able to handle additional inter-cell interference. This is because adaptive filtering is performed independently for each device, using only its training samples, so that the origin of interference to be canceled does not matter. 
We present the average performance of a single cluster. We assume perfect channel estimation for the MMSE-SIC receiver. On the other hand, the PLAF and NLAF do not have channel knowledge and the channel is learned implicitly during the training period. 
Note that it is important that devices are separated well in the SNR domain at the BS for proper function of the MMSE-SIC. The SNR values and other simulation parameters are provided in Table 1. The coherence block length $T_b\leq T_{c}B_c$, where $T_c$ is the coherence time and $B_c$ is the coherence bandwidth, is the number of (complex) symbols a BS can receive before the channel changes to a new independent random value (under the typical assumption of Rayleigh block fading) \cite{Du2017}. All devices in the cluster transmit their training samples simultaneously to the BS. We train the filter for $T$ complex samples (2T real samples) using Algorithm $1$ so there are $T_b-T$ complex data symbols for data transmission.

\begin{table}
\scriptsize
\caption{Parameter Values}
\centering
  \begin{tabular}{| l | l | l |}
    \hline
		 \textbf{Parameter}                  &  \textbf{Symbol }     &   \textbf{Value }\\
		 \hline
		 Number of BS Antennas               &  $M$         &   $3$                      \\
		 Cluster Sizes                        &  $K$         &   $5 \backslash 4 \backslash 3$\\
		 Device SNR                          &  SNR         &   $0,5,10,15,20$ dB         \\
		 Device Spatial Location             &  $\theta$    &   $30^{\circ},60^{\circ},90^{\circ},120^{\circ},150^{\circ}$\\
		 Mobility(Coherence Time)            &  $T_c$       &   $3$-$20$km/h($5$-$15$ms) @$2$GHz See \cite{Sesia2011}\\   
		 Coherence Bandwidth                 &  $B_c$       &   $1$ MHz                  \\
		 Modulation                          &  $b(t)$      &   QAM $[\pm1 \pm i1]$      \\
		 Prob. of Active Devices             &  $\rho$      &   $1 \backslash 0.75 \backslash 0.60$            \\
		 Training Block Size                 &  $T$         &   $500$  \\
		 Dictionary Novelty                  &  $\alpha$    &   $0.1$                    \\
		 Window Size                         &  $W$         &   $50$                     \\
		 Precision                           &  $\epsilon$  &   $0.01$                   \\
		 Gaussian/Linear Weight              &  $w_G, w_L$   &   $0.8, 0.2$                \\ 				
		\hline
 \end{tabular}
\end{table}

\begin{figure}[t]
\centering
    \includegraphics[width=0.41\textwidth]{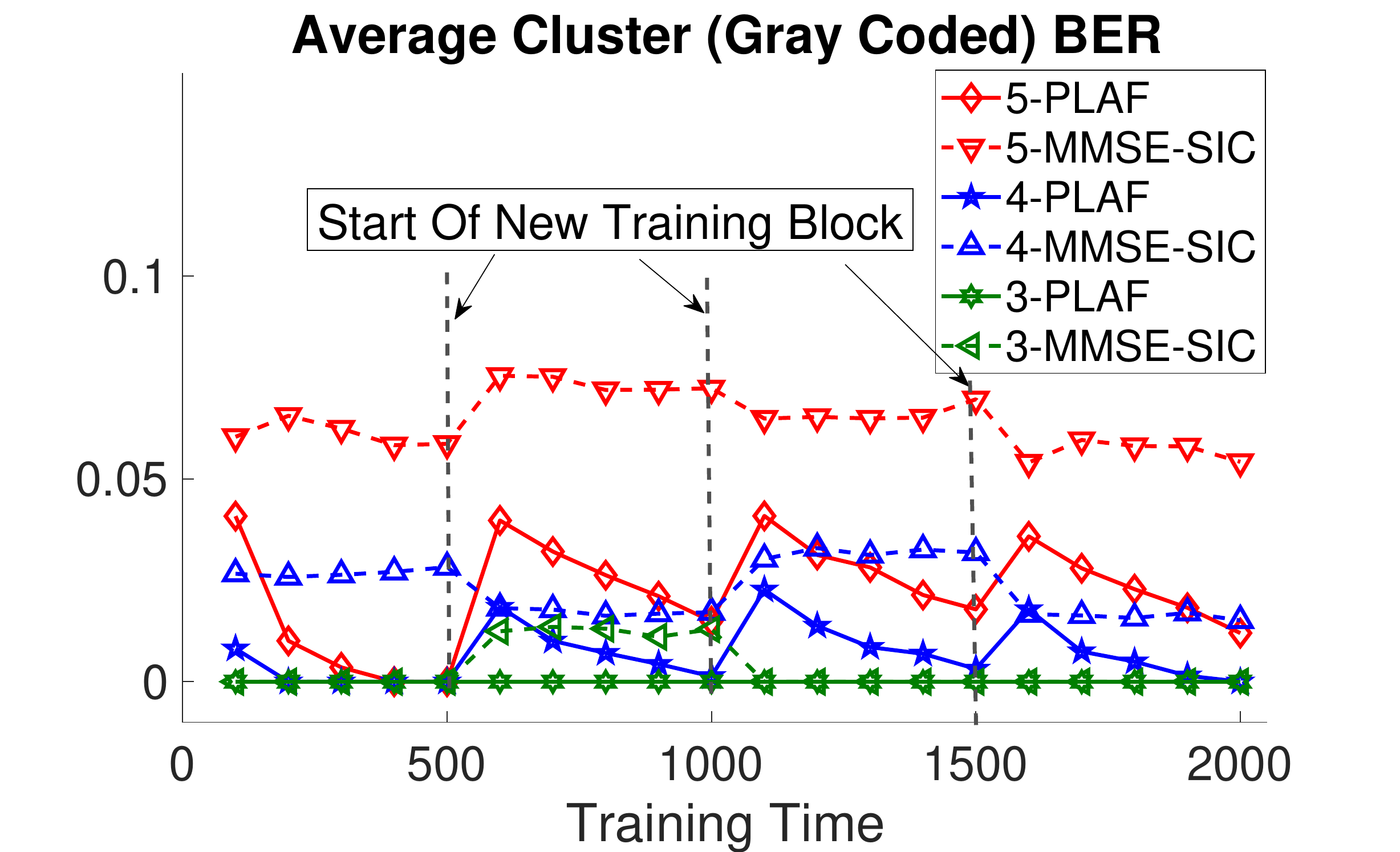}
		\caption{Average BER for PLAF and MMSE-SIC for $K=5 \backslash 4 \backslash 3$, $\rho=0.75$.}
		\label{fig:sum_mmse_comparison}
\end{figure}

\begin{figure}[h!]
\centering
    \includegraphics[width=0.41\textwidth]{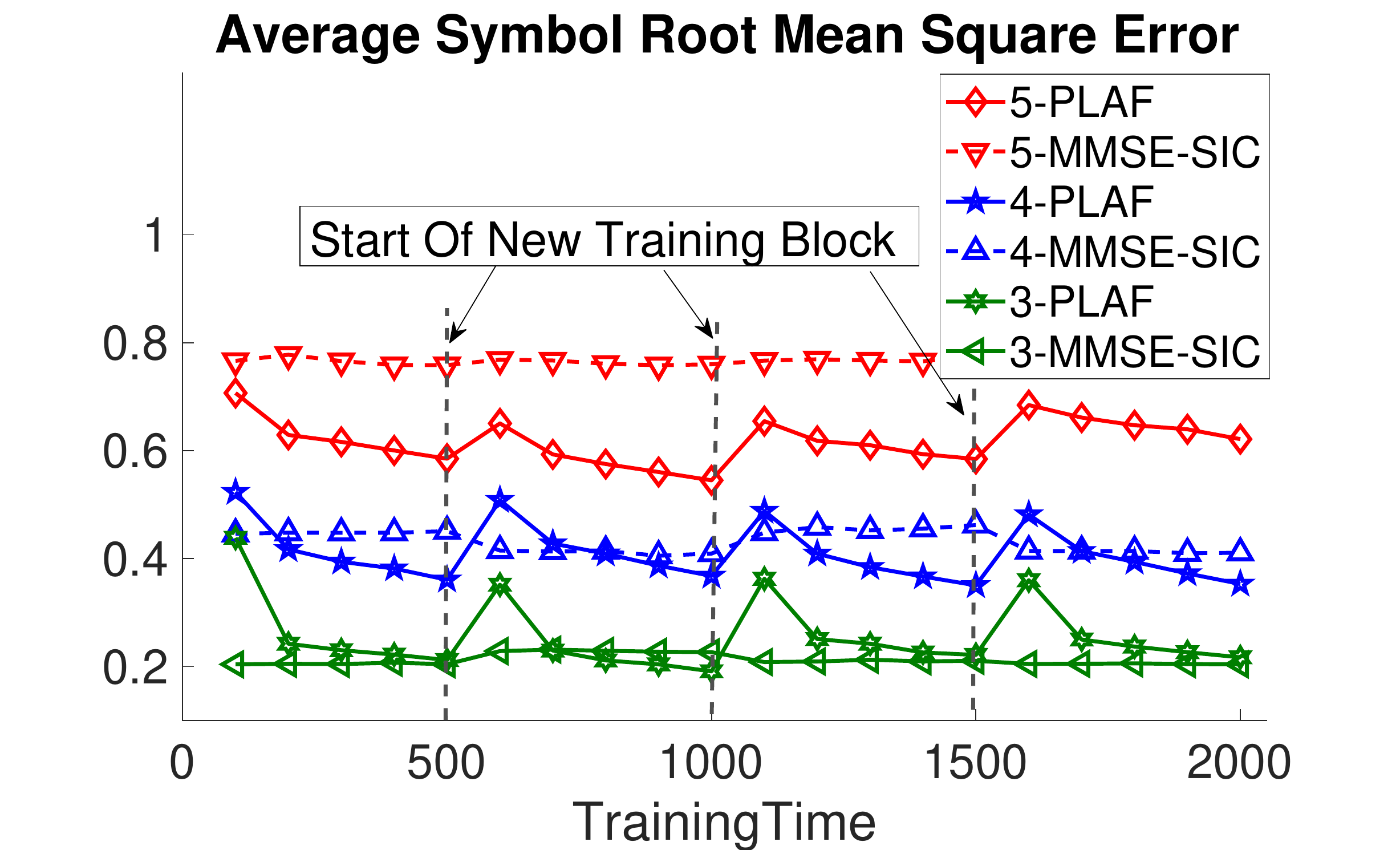}
		\caption{Average cluster RMSE for PLAF and MMSE-SIC for $K=5 \backslash 4 \backslash 3$, $\rho=0.75$.}
		\label{fig:sum_mmse__rms_comparison}
\end{figure}
\begin{figure}[t]
\centering
    \includegraphics[width=0.41\textwidth]{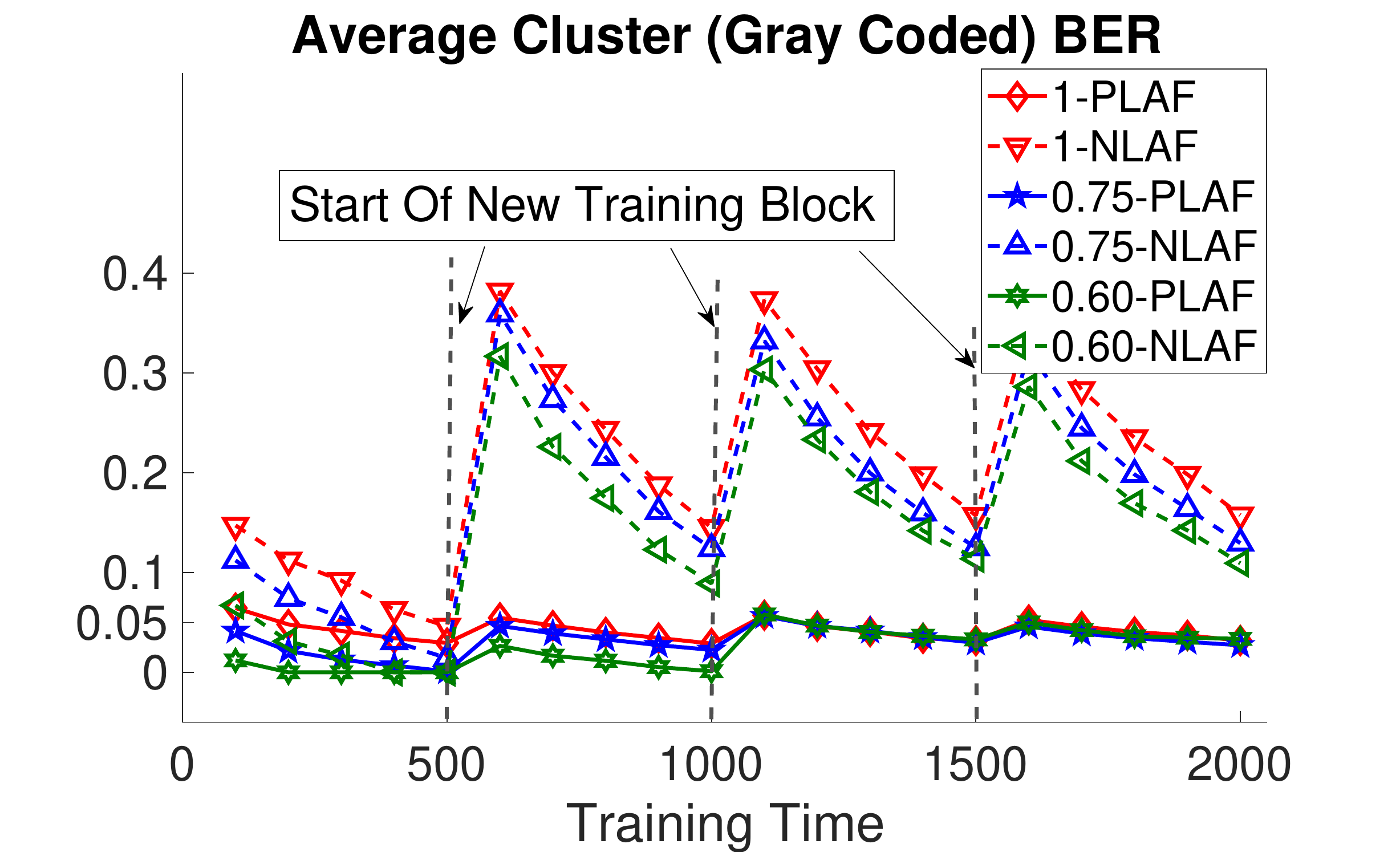}
		\caption{Average BER for PLAF and NLAF for $K=5$, $\rho=1 \backslash 0.75 \backslash 0.60$.}
		\label{fig:sum_rbf_comparison}
\end{figure}
\begin{figure}[h!]
\centering
    \includegraphics[width=0.41\textwidth]{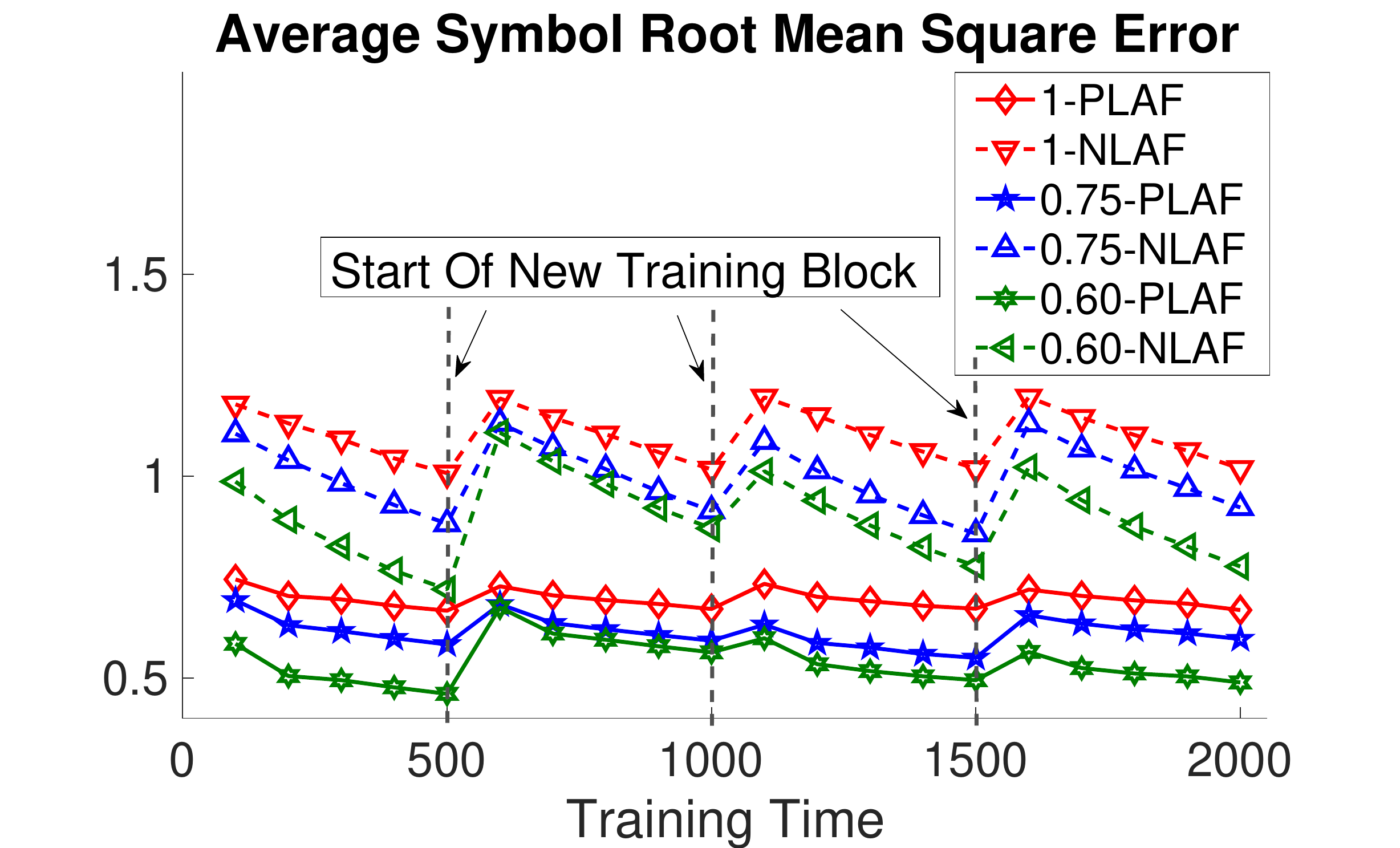}
		\caption{Average cluster RMSE for PLAF and NLAF for $K=5$, $\rho=1 \backslash 0.75 \backslash 0.60$.}
		\label{fig:sum_rbf_rms_comparison}
\end{figure}
\subsection{Pruning Strategy}\label{sec:pruning_strategy} 
The maximum size of the Gaussian dictionary is chosen to be $3T$. During training block $T$ in Algorithm $1$, the size of the dictionary can at most grow by $2T$. We start with $D_{G,-1}=\emptyset$. At the end of each training block, we only keep the most recent $T$ elements and drop the rest to maintain the size $S_n:=|D_{G,n}|\leq3T$. At the start of the next training, $D_{G,-1}$ contains the last $T$ elements from the previous block. This strategy is reasonable since the change in channel conditions and device activity at the start of each training block makes the old filter coefficients less relevant, so we can discard these to admit new samples in the next block.        

\subsection{Results}
We compare the performance of PLAF, NLAF and MMSE-SIC in terms of average gray coded bit error rate (BER) and \textit{root mean-square error} (RMSE) for $4$ successive training blocks using Algorithm $1$. Each block is $T=500$ symbols long. SIC is performed on the symbol level, i.e., each modulated symbol from the interfering device is detected and then canceled without decoding. This technique is known as \textit{symbol level interference cancellation} (SLIC) which has a lower implementation complexity compared to the code level SIC \cite{Saito2015}. For simulation, at the start of each training block/period (marked by a dashed vertical line) the Rayleigh fading channel is changed to a new random independent value and a new set of active devices is selected randomly. The scenario models the dynamic environment of, for example, 5G IoT systems. The learning model is validated in intervals of $100$ training symbols. We used $300$ test data points to calculate the average gray coded bit error rate (BER) by hard-decision detection and the RMSE. All results are a uniform average of $100$ experiments. 

In Figures \ref{fig:sum_mmse_comparison} and \ref{fig:sum_mmse__rms_comparison} we compare the performance of the proposed PLAF and MMSE-SIC based detection for cluster sizes $K=3$, $K=4$, and $K=5$. For this experiment we fix the probability of all devices transmitting in a coherence block at $\rho=0.75$. We notice that with increased cluster sizes of $K>M$ the performance of MMSE-SIC despite of perfect channel information deteriorates while PLAF shows a robust superior performance. This behavior is met because MMSE-SIC suffers from too much interference and errors in detection are propagated through the SIC chain. 

In Figures \ref{fig:sum_rbf_comparison} and  \ref{fig:sum_rbf_rms_comparison} we compare the performance of the proposed PLAF and NLAF for $\rho=1$, $\rho=0.75$, and $\rho=0.60$. We fix the cluster size to $K=5$. We observe that a change in the channel conditions and active device distribution causes much deterioration in the performance of NLAF which is very sensitive to changes in environment. The PLAF on the other hand, shows robustness against these variations.

\section*{Acknowledgment}
This work has been performed in the framework of the Horizon 2020 project ONE5G (ICT-760809) receiving funds from the European Union. The authors would like to acknowledge the contributions of their colleagues in the project, although the views expressed in this contribution are those of the authors and do not necessarily represent the project. This research was also supported by Grant STA 864/9-1 from German Research Foundation (DFG) and JSPS Grants-in-Aid (15K06081).

\appendix
\subsection{Calculation of $\mathbf{P}_{\mathcal{H}_{G,n-1}}(\kappa_{G}(\mathbf{r}_n,\cdot))$ and $d_n$}\label{app:A}
$\mathbf{P}_{\mathcal{H}_{G,n-1}}(\kappa_{G}(\mathbf{r}_n,\cdot))$ is defined as \cite{SlavakisSliding2008}
\begin{equation}
\mathbf{P}_{\mathcal{H}_{G,n-1}}(\kappa(\mathbf{r}_{n},\cdot)):=\sum_{l=1}^{S_{n-1}}\boldsymbol{\zeta}_{\mathbf{r}_n,l}^{n}\boldsymbol{\Psi}^{n-1}_{l},
\label{eqn:projection_one}
\end{equation}
where $S_{n-1}:=|\mathcal{D}_{G,n-1}|$ is the cardinality of $\mathcal{D}_{G,n-1}$, $\mathbb{R}^{S_{n-1}}\ni\boldsymbol{\zeta}_{\mathbf{r}_n}^{n}=\mathbf{K}_{n-1}^{-1}\boldsymbol{\xi}^{n}_{\mathbf{r}_n}$ and $\boldsymbol{\Psi}^{n-1}_{l}\in\mathcal{D}_{G,n-1}$ is the $l$th element of the Gaussian dictionary up to time $n-1$.  With $\mathbf{K}_{0}^{-1}:=1/\kappa(\mathbf{r}_0,\mathbf{r}_0)$, $\mathbf{K}_{n}^{-1}$ is  given by the recursion
\[
\mathbf{K}_{n}^{-1}:=
\begin{bmatrix}
    \mathbf{K}_{n-1}^{-1}+\frac{\boldsymbol{\zeta}_{\mathbf{r}_n}^{n}(\boldsymbol{\zeta}_{\mathbf{r}_n}^{n})^{t}}{d_n^{2}}   & -\frac{\boldsymbol{\zeta}_{\mathbf{r}_n}^{n}}{d_n^{2}} \\
    -\frac{(\boldsymbol{\zeta}_{\mathbf{r}_n}^{n})^{t}}{d_n^{2}}                                                           &  \frac{1}{d_n^{2}}   
\end{bmatrix}
\]
if $\kappa_G(\mathbf{r}_n,\cdot)\in\mathcal{D}_{G,n}$, otherwise $\mathbf{K}_{n}^{-1}:=\mathbf{K}_{n-1}^{-1}$, and 
\[
     \boldsymbol{\xi}^{n}_{\mathbf{r}_n}:= 
		\begin{bmatrix}
		 \left\langle \kappa_{G}(\mathbf{r}_n,\cdot),\boldsymbol{\Psi}^{n-1}_{1}\right\rangle_{\mathcal{H}_G}\\
		\vdots\\
		 \left\langle \kappa_{G}(\mathbf{r}_n,\cdot),\boldsymbol{\Psi}^{n-1}_{S_{n-1}}\right\rangle_{\mathcal{H}_G}
		\end{bmatrix}\in\mathbb{R}^{S_{n-1}}
\]

The distance $d_n$ is given by $d_n:=\kappa_G(\mathbf{r}_n,\mathbf{r}_n)-(\boldsymbol{\xi}^{n}_{\mathbf{r}_n})^{t}\boldsymbol{\zeta}_{\mathbf{r}_n}^{n}$ \cite{SlavakisSliding2008}.
 \subsection{Calculation of $\mathbf{P}_{\mathcal{H}_{G,n}}(\kappa(\mathbf{r}_j,\cdot))$} \label{app:B}
First consider $j=n$. If $\kappa_G(\mathbf{r}_n,\cdot)\in\mathcal{D}_{G,n}$, $\mathbf{P}_{\mathcal{H}_{G,n}}(\kappa_G(\mathbf{r}_n,\cdot))=\kappa_G(\mathbf{r}_n,\cdot)$. Otherwise $\mathbf{P}_{\mathcal{H}_{G,n}}(\kappa_G(\mathbf{r}_n,\cdot))=\mathbf{P}_{\mathcal{H}_{G,n-1}}(\kappa_G(\mathbf{r}_n,\cdot))$. But $\mathbf{P}_{\mathcal{H}_{G,n-1}}(\kappa_G(\mathbf{r}_j,\cdot))$ is available to us by the \textit{Dictionary Update Step}. Since $\mathcal{H}_{G,n-1}\subseteq\mathcal{H}_{G,n}$ for each $n$, it follows that $\mathbf{P}_{\mathcal{H}_{G,n}}(\kappa_G(\mathbf{r}_j,\cdot))$ is available to us $\forall j\in\mathcal{J}_n$. 

\subsection{Calculation of $\beta_{j}^{n}$} \label{app:C}
For each $j\in\mathcal{J}_n$, \cite{Slavakis2009}
\[ 
\beta^{n}_{j}:=\Bigg\{\begin{tabular}{cc}
  $\frac{b_j-\left\langle f_n,\kappa(\mathbf{r}_j,\cdot)\right\rangle_{\mathcal{H}}-\epsilon}{\kappa(\mathbf{r}_j,\mathbf{r}_j)},$ & if  $\left\langle f_n,\kappa(\mathbf{r}_j,\cdot)\right\rangle_{\mathcal{H}}-b_j < -\epsilon$\\
  0, &\, if $|\left\langle f_n,\kappa(\mathbf{r}_j,\cdot)\right\rangle_{\mathcal{H}}-b_j| \leq \epsilon$\\
  $\frac{b_j-\left\langle f_n,\kappa(\mathbf{r}_j,\cdot)\right\rangle_{\mathcal{H}}+\epsilon}{\kappa(\mathbf{r}_j,\mathbf{r}_j)},$ & if  $\left\langle f_n,\kappa(\mathbf{r}_j,\cdot)\right\rangle_{\mathcal{H}}-b_j > \epsilon$
  \end{tabular}
\]
where $\epsilon$ is a design parameter.


\bibliographystyle{IEEEtran}
\bibliography{library}

\end{document}